\documentclass{article}
\usepackage{nips14submit_e,times}
\usepackage{hyperref}
\usepackage{url}

\usepackage{graphicx}
\usepackage{subfig}

\usepackage{algorithm}
\usepackage{algorithmic}
\usepackage{amsfonts}
\usepackage{amsmath}

\usepackage{booktabs}

\title{Deep Speech:  Scaling up end-to-end \\ speech recognition}

\author{Awni Hannun\thanks{Contact author:  awnihannun@baidu.com}, Carl Case, Jared Casper, Bryan Catanzaro, Greg Diamos, Erich Elsen, \And Ryan Prenger, Sanjeev Satheesh, Shubho Sengupta, Adam Coates, Andrew Y. Ng \\[10pt]
Baidu Research -- Silicon Valley AI Lab
}

\nipsfinalcopy

\begin{document} 
\maketitle
\vskip -0.2in
\begin{abstract} 

We present a state-of-the-art speech recognition system developed using
end-to-end deep learning.  Our architecture is significantly simpler than
traditional speech systems, which rely on laboriously engineered processing
pipelines; these traditional systems also tend to perform poorly when used in
noisy environments.  In contrast, our system does not need hand-designed
components to model background noise, reverberation, or speaker variation, but
instead directly learns a function that is robust to such effects.  We do not
need a phoneme dictionary, nor even the concept of a ``phoneme.''  Key to our
approach is a well-optimized RNN training system that uses multiple GPUs, as
well as a set of novel data synthesis techniques that allow us to efficiently
obtain a large amount of varied data for training.  Our system, called Deep
Speech, outperforms previously published results on the widely studied
Switchboard Hub5'00, achieving 16.0\% error on the full test set.  Deep Speech
also handles challenging noisy environments better than widely used,
state-of-the-art commercial speech systems. 

\end{abstract}

\section{Introduction}
\label{sec:introduction}

Top speech recognition systems rely on sophisticated pipelines composed of
multiple algorithms and hand-engineered processing stages.  In this paper, we
describe an end-to-end speech system, called ``Deep Speech'', where deep
learning supersedes these processing stages.  Combined with a language model,
this approach achieves higher performance than traditional methods on hard
speech recognition tasks while also being much simpler.  These results are made
possible by training a large recurrent neural network (RNN) using multiple GPUs
and thousands of hours of data.  Because this system learns directly from data,
we do not require specialized components for speaker adaptation or noise
filtering.  In fact, in settings where robustness to speaker variation and
noise are critical, our system excels:  Deep Speech outperforms previously
published methods on the Switchboard Hub5'00 corpus, achieving 16.0\% error,
and performs better than commercial systems in noisy speech recognition tests.

Traditional speech systems use many heavily engineered processing stages,
including specialized input features, acoustic models, and Hidden Markov Models
(HMMs).  To improve these pipelines, domain experts must invest a great deal of
effort tuning their features and models.  The introduction of deep learning
algorithms~\cite{lee2009cdbn_audio,Mohamed2011,grosse2012shift,Hinton2012,Dahl2011a}
has improved speech system performance, usually by improving acoustic models.
While this improvement has  been significant, deep learning still plays only a
limited role in traditional speech pipelines.  As a result, to improve
performance on a task such as recognizing speech in a noisy environment, one
must laboriously engineer the rest of the system for robustness.  In contrast,
our system applies deep learning end-to-end using recurrent neural networks.
We take advantage of the capacity provided by deep learning systems to learn
from large datasets to improve our overall performance.  Our model is trained
end-to-end to produce transcriptions and thus, with sufficient data and
computing power, can learn robustness to noise or speaker variation on its own.

Tapping the benefits of end-to-end deep learning, however, poses several
challenges:  (i) we must find innovative ways to build large, labeled training
sets and (ii) we must be able to train networks that are large enough to
effectively utilize all of this data.  One challenge for handling labeled data
in speech systems is finding the alignment of text transcripts with input
speech.  This problem has been addressed by Graves, Fern\'{a}ndez, Gomez and
Schmidhuber~\cite{Graves2006}, thus enabling neural networks to easily consume
unaligned, transcribed audio during training.  Meanwhile, rapid training of
large neural networks has been tackled by Coates et
al.~\cite{coates2013cotshpc}, demonstrating the speed advantages of multi-GPU
computation.  We aim to leverage these insights to fulfill the vision of a
generic learning system, based on large speech datasets and scalable RNN
training, that can surpass more complicated traditional methods.  This vision
is inspired partly by the work of Lee~et.~al.~\cite{lee2009cdbn_audio} who
applied early unsupervised feature learning techniques to replace hand-built
speech features.

We have chosen our RNN model specifically to map well to GPUs and we use a
novel model partition scheme to improve parallelization.  Additionally, we
propose a process for assembling large quantities of labeled speech data
exhibiting the distortions that our system should learn to handle.  Using a
combination of collected and synthesized data, our system learns robustness to
realistic noise and speaker variation (including Lombard
Effect~\cite{junqua1993lombard}).  Taken together, these ideas suffice to build
an end-to-end speech system that is at once simpler than traditional pipelines
yet also performs better on difficult speech tasks.  Deep Speech achieves an
error rate of 16.0\% on the full Switchboard Hub5'00 test set---the best
published result.  Further, on a new noisy speech recognition dataset of our
own construction, our system achieves a word error rate of 19.1\% where the
best commercial systems achieve 30.5\% error.

In the remainder of this paper, we will introduce the key ideas behind our
speech recognition system.  We begin by describing the basic recurrent neural
network model and training framework that we use in
Section~\ref{section:model}, followed by  a discussion of GPU optimizations
(Section~\ref{section:optimization}), and our data capture and synthesis
strategy (Section~\ref{section:data}).  We conclude with our experimental
results demonstrating the state-of-the-art performance of Deep Speech
(Section~\ref{section:experiments}), followed by a discussion of related work
and our conclusions.

\section{RNN Training Setup}
\label{section:model}
The core of our system is a recurrent neural network (RNN) trained to ingest
speech spectrograms and generate English text transcriptions.  Let a single
utterance $x$ and label $y$ be sampled from a training set $\mathcal{X} =
\{(x^{(1)},y^{(1)}),(x^{(2)},y^{(2)}),\ldots\}$.  Each utterance, $x^{(i)}$, is
a time-series of length $T^{(i)}$ where every time-slice is a vector of audio
features, $x_t^{(i)}, t=1,\ldots,T^{(i)}$.  We use spectrograms as our
features, so $x^{(i)}_{t,p}$ denotes the power of the $p$'th frequency bin in
the audio frame at time $t$.  The goal of our RNN is to convert an input
sequence $x$ into a sequence of character probabilities for the transcription
$y$, with $\hat{y_t} = \mathbb{P}(c_t|x)$, where $c_t \in
\{\textrm{a,b,c,}\ldots,\textrm{z},
\textit{space},\textit{apostrophe},\textit{blank}\}$.

Our RNN model is composed of 5 layers of hidden units.  For an input $x$, the
hidden units at layer $l$ are denoted $h^{(l)}$ with the convention that
$h^{(0)}$ is the input.  The first three layers are not recurrent.  For the
first layer, at each time $t$, the output depends on the spectrogram frame
$x_t$ along with a context of $C$ frames on each side.\footnote{We typically
use $C\in \{5, 7, 9\}$ for our experiments.} The remaining non-recurrent layers
operate on independent data for each time step.
    Thus, for each time $t$, the first 3 layers are computed by:
        \begin{align*}
            h^{(l)}_t &= g(W^{(l)} h^{(l-1)}_t + b^{(l)})
        \end{align*}
where $g(z) = \min\{\max\{0,z\}, 20\}$ is the clipped rectified-linear (ReLu)
activation function and $W^{(l)}, b^{(l)}$ are the weight matrix and bias
parameters for layer $l$.\footnote{The ReLu units are clipped in order to keep
the activations in the recurrent layer from exploding; in practice the units
rarely saturate at the upper bound.} The fourth layer is a bi-directional
recurrent layer~\cite{schuster1997bidirectional}.  This layer includes two sets
of hidden units:  a set with forward recurrence, $h^{(f)}$, and a set with
backward recurrence $h^{(b)}$:
    \begin{align*}
    h^{(f)}_t &= g(W^{(4)} h^{(3)}_t + W_r^{(f)} h^{(f)}_{t-1} + b^{(4)}) \\
    h^{(b)}_t &= g(W^{(4)} h^{(3)}_t + W_r^{(b)} h^{(b)}_{t+1} + b^{(4)})
    \end{align*}
Note that $h^{(f)}$ must be computed sequentially from $t=1$ to $t=T^{(i)}$ for
the $i$'th utterance, while the units $h^{(b)}$ must be computed sequentially
in reverse from $t=T^{(i)}$ to $t=1$.

The fifth (non-recurrent) layer takes both the forward and backward units as
inputs $h^{(5)}_t = g(W^{(5)} h^{(4)}_t + b^{(5)})$ where $h^{(4)}_t =
h^{(f)}_t + h^{(b)}_t$.  The output layer is a standard softmax function that
yields the predicted character probabilities for each time slice $t$ and
character $k$ in the alphabet:
\begin{align*}
h_{t,k}^{(6)} = \hat{y}_{t,k} \equiv \mathbb{P}(c_t = k|x) =  \frac{\exp(W_k^{(6)} h_t^{(5)}+b_k^{(6)})}{\sum_j \exp(W_j^{(6)} h_t^{(5)}+b_j^{(6)})}.
\end{align*}
Here $W_k^{(6)}$ and $b_k^{(6)}$ denote the $k$'th column of the weight matrix
and $k$'th bias, respectively.  

Once we have computed a prediction for $\mathbb{P}(c_t|x)$, we compute the CTC
loss~\cite{Graves2006} $\mathcal{L}(\hat{y}, y)$ to measure the error in
prediction.  During training, we can evaluate the gradient $\nabla_{\hat{y}}
\mathcal{L}(\hat{y}, y)$ with respect to the network outputs given the
ground-truth character sequence $y$.  From this point, computing the gradient
with respect to all of the model parameters may be done via back-propagation
through the rest of the network. We use Nesterov's Accelerated gradient method
for training~\cite{sutskever2013nag}.\footnote{We use momentum of 0.99 and
anneal the learning rate by a constant factor, chosen to yield the fastest
convergence, after each epoch through the data.}

\begin{figure}[th]
\centering
 \includegraphics[width=0.6\textwidth]{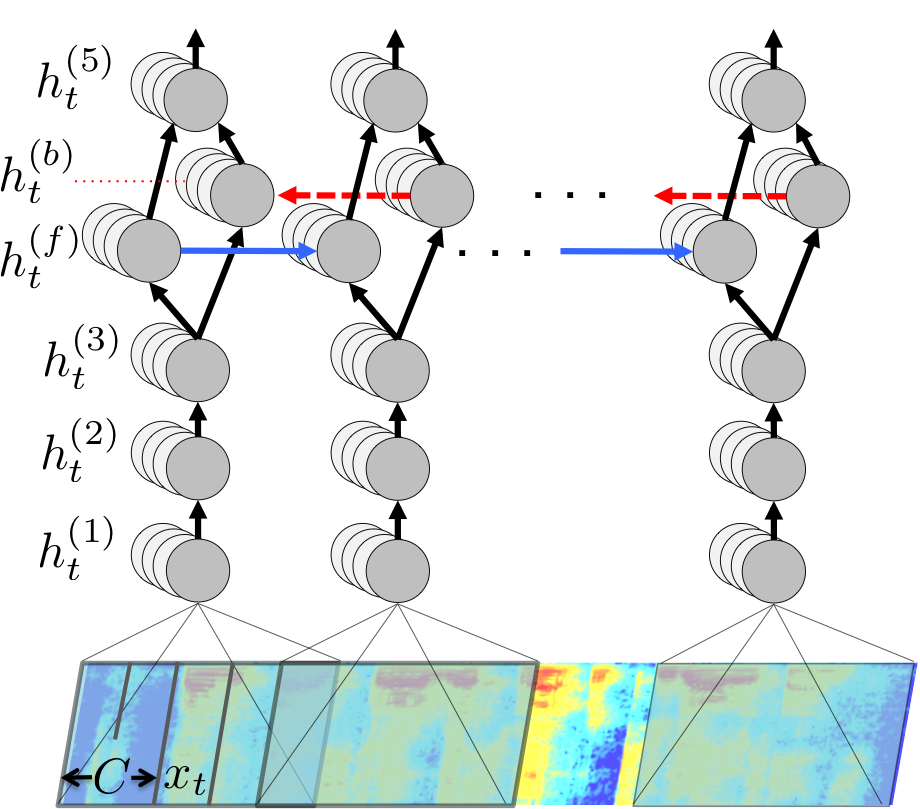}
  \caption{Structure of our RNN model and notation.}
  \label{fig:rnn}
\end{figure}

The complete RNN model is illustrated in Figure~\ref{fig:rnn}.  Note that its
structure is considerably simpler than related models from the
literature~\cite{Graves2014}---we have limited ourselves to a single recurrent
layer (which is the hardest to parallelize) and we do not use
Long-Short-Term-Memory (LSTM) circuits.  One disadvantage of LSTM cells is that
they require computing and storing multiple gating neuron responses at each
step.  Since the forward and backward recurrences are sequential, this small
additional cost can become a computational bottleneck.  By using a homogeneous
model we have made the computation of the recurrent activations as efficient as
possible:  computing the ReLu outputs involves only a few highly optimized BLAS
operations on the GPU and a single point-wise nonlinearity.

\subsection{Regularization}
While we have gone to significant lengths to expand our datasets (c.f.
Section~\ref{section:data}), the recurrent networks we use are still adept at
fitting the training data.  In order to reduce variance further, we use several
techniques.  

During training we apply a dropout~\cite{hinton2012dropout} rate between 5\% -
10\%. We apply dropout in the feed-forward layers but not to the recurrent
hidden activations.

A commonly employed technique in computer vision during network evaluation is
to randomly jitter inputs by translations or reflections, feed each jittered
version through the network, and vote or average the
results~\cite{krizhevsky2012imagenet}. Such jittering is not common in ASR,
however we found it beneficial to translate the raw audio files by 5ms (half
the filter bank step size) to the left and right, then forward propagate the
recomputed features and average the output probabilities. At test time we also
use an ensemble of several RNNs, averaging their outputs in the same way.

\subsection{Language Model}
\label{section:languagemodel}

When trained from large quantities of labeled speech data, the RNN model can
learn to produce readable character-level transcriptions.  Indeed for many of
the transcriptions, the most likely character sequence predicted by the RNN is
exactly correct without external language constraints.  The errors made by the
RNN in this case tend to be phonetically plausible renderings of English
words---Table~\ref{table:max_decoded} shows some examples.  Many of the errors
occur on words that rarely or never appear in our training set.  In practice,
this is hard to avoid:  training from enough speech data to \emph{hear} all of
the words or language constructions we might need to know is impractical.
Therefore, we integrate our system with an N-gram language model since these
models are easily trained from huge unlabeled text corpora.  For comparison,
while our speech datasets typically include up to 3 million utterances, the
N-gram language model used for the experiments in
Section~\ref{section:expnoise} is trained from a corpus of 220 million phrases,
supporting a vocabulary of 495,000 words.\footnote{We use the KenLM
toolkit~\cite{heafield2013kenlm} to train the N-gram language models in our
experiments.} 

\begin{table}[h]
\centering
\begin{tabular}{l | l}
\toprule
RNN output  & Decoded Transcription   \\
\midrule
what is the weather like in bostin right now & what is the weather like in boston right now \\
prime miniter nerenr modi   & prime minister narendra modi     \\ 
arther n tickets for the game  & are there any tickets for the game        \\ 
\bottomrule
\end{tabular}
\caption{Examples of transcriptions directly from the RNN (left) with errors that are fixed by addition of a language model (right).}
\label{table:max_decoded}
\end{table}

Given the output $\mathbb{P}(c|x)$ of our RNN we perform a search to find the sequence of characters $c_1,c_2,\ldots$ that is most probable according to both the RNN output and the language model (where the language model interprets the string of characters as words). Specifically, we aim to find a sequence $c$ that maximizes the combined objective:
\begin{align*}
Q(c) = \log(\mathbb{P}(c|x)) + \alpha \log(\mathbb{P}_{\text{lm}}(c)) + \beta \textrm{ word\_count}(c)
\end{align*}
where $\alpha$ and $\beta$ are tunable parameters (set by cross-validation) that control the trade-off between the RNN, the language model constraint and the length of the sentence.  The term $\mathbb{P}_{\text{lm}}$ denotes the probability of the sequence $c$ according to the N-gram model.  We maximize this objective using a highly optimized beam search algorithm, with a typical beam size in the range 1000-8000---similar to the approach described by Hannun et al.~\cite{hannun2014firstpass}.

\section{Optimizations}
\label{section:optimization}

As noted above, we have made several design decisions to make
our networks amenable to high-speed execution (and thus fast training).
For example, we have opted for homogeneous rectified-linear networks
that are simple to implement and depend on just a few highly-optimized BLAS
calls.  When fully unrolled, our networks include almost 5 billion connections
for a typical utterance and thus efficient computation is critical to 
make our experiments feasible.  We use multi-GPU 
training~\cite{coates2013cotshpc,krizhevsky2012imagenet} to accelerate our 
experiments, but doing this effectively requires some additional work, as we 
explain.

\subsection{Data parallelism}
\label{section:datapar}

In order to process data efficiently, we use two levels of data parallelism.  First, each GPU processes many examples in parallel.  This is done in the usual way by concatenating many examples into a single matrix.  For instance, rather than performing a single matrix-vector multiplication $W_r h_t$ in the recurrent layer, we prefer to do many in parallel by computing $W_r H_t$ where $H_t = [ h^{(i)}_{t}, h^{(i+1)}_{t}, \ldots ]$ 
(where $h_t^{(i)}$ corresponds to the $i$'th example $x^{(i)}$ at time $t$).  The GPU is most
efficient when $H_t$ is relatively wide (e.g., 1000 examples or more) and thus we prefer to process as many
examples on one GPU as possible (up to the limit of GPU memory).

When we wish to use larger minibatches than a single GPU can support on its own
we use data parallelism across multiple GPUs, with each GPU processing a
separate minibatch of examples and then combining its computed gradient with
its peers during each iteration.  We typically use $2\times$ or $4\times$ data
parallelism across GPUs.

Data parallelism is not easily implemented, however, when utterances have
different lengths since they cannot be combined into a single matrix
multiplication.  We resolve the problem by sorting our training examples by
length and combining only similarly-sized utterances into minibatches, padding
with silence when necessary so that all utterances in a batch have the same
length.  This solution is inspired by the ITPACK/ELLPACK sparse matrix
format~\cite{Kincaid:89};  a similar solution was used by the Sutskever et
al.~\cite{sutskever2014seq} to accelerate RNNs for text.

\subsection{Model parallelism}

Data parallelism yields training speedups for modest multiples of the minibatch
size (e.g., 2 to 4), but faces diminishing returns as batching more examples
into a single gradient update fails to improve the training convergence rate.
That is, processing $2\times$ as many examples on $2\times$ as many GPUs fails
to yield a $2\times$ speedup in training.  It is also inefficient to fix the
total minibatch size but spread out the examples to $2\times$ as many GPUs:  as
the minibatch \emph{within} each GPU shrinks, most operations become
memory-bandwidth limited.  To scale further, we parallelize by partitioning the
model (``model parallelism''~\cite{coates2013cotshpc,dean2012largescale}).

Our model is challenging to parallelize due to the sequential nature of the
recurrent layers.  Since the bidirectional layer is comprised of a forward
computation and a backward computation that are independent, we can perform the
two computations in parallel.  Unfortunately, naively splitting the RNN to
place $h^{(f)}$ and $h^{(b)}$ on separate GPUs commits us to significant data
transfers when we go to compute $h^{(5)}$ (which depends on both $h^{(f)}$ and
$h^{(b)}$).  Thus, we have chosen a different partitioning of work that
requires less communication for our models:  we divide the model in half along
the \emph{time} dimension.

All layers except the recurrent layer can be trivially decomposed along the
time dimension, with the first half of the time-series, from $t=1$ to
$t=T^{(i)}/2$, assigned to one GPU and the second half to another GPU.  When
computing the recurrent layer activations, the first GPU begins computing the
forward activations $h^{(f)}$, while the second begins computing the backward
activations $h^{(b)}$.  At the mid-point ($t=T^{(i)}/2$), the two GPUs exchange
the intermediate activations, $h^{(f)}_{T/2}$ and $h^{(b)}_{T/2}$ and swap
roles.  The first GPU then finishes the backward computation of $h^{(b)}$ and
the second GPU finishes the forward computation of $h^{(f)}$.

\subsection{Striding}

We have worked to minimize the running time of the recurrent layers of our RNN,
since these are the hardest to parallelize.  As a final optimization, we
shorten the recurrent layers by taking ``steps'' (or strides) of size 2 in the
original input so that the unrolled RNN has half as many steps.  This is
similar to a convolutional network~\cite{lecun1989convnet} with a step-size of
2 in the first layer.  We use the cuDNN library~\cite{Chetlur:14} to implement
this first layer of convolution efficiently.

\section{Training Data}
\label{section:data}

Large-scale deep learning systems require an abundance of labeled data.  For
our system we need many recorded utterances and corresponding English
transcriptions, but there are few public datasets of sufficient scale.  To
train our largest models we have thus collected an extensive dataset consisting
of 5000 hours of read speech from 9600 speakers.  For comparison, we have
summarized the labeled datasets available to us in Table~\ref{table:datasets}.

\begin{table}[]
\centering
\begin{tabular}{l c c c}
 \toprule
 Dataset & Type & Hours & Speakers  \\
 \midrule
 WSJ         & read           &   80 & 280 \\
 Switchboard & conversational &  300 & 4000 \\
 Fisher      & conversational & 2000 & 23000 \\
 Baidu       &  read          & 5000 & 9600 \\
 \bottomrule
\end{tabular}
\caption{A summary of the datasets used to train Deep Speech. The Wall Street Journal, Switchboard and Fisher~\cite{Cieri2004Fisher} corpora are all published by the Linguistic Data Consortium.}
\label{table:datasets}
\end{table}

\subsection{Synthesis by superposition}
\label{section:noisesynth}

To expand our potential training data even further we use data synthesis, which
has been successfully applied in other contexts to amplify the effective number
of training
samples~\cite{sapp2008synth,lecun2004learningmethods,coates2011icdar}. In our
work, the goal is primarily to improve performance in noisy environments where
existing systems break down.  Capturing labeled data (e.g., read speech) from
noisy environments is not practical, however, and thus we must find other ways
to generate such data.

To a first order, audio signals are generated through a process of
superposition of source signals.  We can use this fact to synthesize noisy
training data.  For example, if we have a speech audio track $x^{(i)}$ and a
``noise'' audio track $\xi^{(i)}$, then we can form the ``noisy speech'' track
$\hat{x}^{(i)} = x^{(i)}+\xi^{(i)}$ to simulate audio captured in a noisy
environment.  If necessary, we can add reverberations, echoes or other forms of
damping to the power spectrum of $\xi^{(i)}$ or $x^{(i)}$ and then simply add
them together to make fairly realistic audio scenes.

There are, however, some risks in this approach. For example, in order
to take 1000 hours of clean speech and create 1000 hours of noisy speech,
we will need unique noise tracks spanning roughly 1000 hours.  
We cannot settle for, say, 10 hours of
repeating noise, since it may become possible for the recurrent
network to memorize the noise track and ``subtract'' it out of the
synthesized data.  Thus, instead of using a single noise source
$\xi^{(i)}$ with a length of 1000 hours, we use a large number
of shorter clips (which are easier to collect from public video
sources) and treat them as separate sources of noise before
superimposing all of them: $\hat{x}^{(i)} = x^{(i)} + \xi_1^{(i)}
+\xi_2^{(i)} + \ldots$.

When superimposing many signals collected from video clips, we can end up with
``noise'' sounds that are different from the kinds of noise recorded in real
environments.  To ensure a good match between our synthetic data and real data,
we rejected any candidate noise clips where the average power in each frequency
band differed significantly from the average power observed in real noisy
recordings.

\subsection{Capturing Lombard Effect}
\label{section:lombard}
One challenging effect encountered by speech recognition systems in noisy
environments is the ``Lombard Effect''~\cite{junqua1993lombard}:  
speakers actively change the pitch or inflections of their voice to overcome noise around them.  This
(involuntary) effect does not show up in recorded speech datasets since
they are collected in quiet environments.  To ensure that the effect is represented in our training data we induce the Lombard effect intentionally during data collection by playing loud background noise through headphones worn by a person as they record an utterance.  The noise induces them to inflect their voice, thus allowing us to capture the Lombard effect in our training data.\footnote{We 
have experimented with noise played through headphones as well as through computer speakers.  Using headphones has the advantage that we obtain ``clean'' recordings
without the background noise included and can add our own synthetic noise afterward.}

\section{Experiments}
\label{section:experiments}

We performed two sets of experiments to evaluate our system.  In both cases we
use the model described in Section~\ref{section:model} trained from a selection
of the datasets in Table~\ref{table:datasets} to predict character-level
transcriptions.  The predicted probability vectors and language model are then
fed into our decoder to yield a word-level transcription, which is compared
with the ground truth transcription to yield the word error rate (WER).

\subsection{Conversational speech:  Switchboard Hub5'00 (full)}
% AN:  take credit for scaling up to 2300 hrs of speech.
% Scaling up expensive/hard;  other teams often use smaller datasets.
%
To compare our system to prior research we use an accepted but highly
challenging test set, Hub5'00 (LDC2002S23).  Some researchers split this set
into ``easy'' (Switchboard) and ``hard'' (CallHome) instances, often reporting
new results on the easier portion alone.  We use the full set, which is the
most challenging case and report the overall word error rate.

We evaluate our system trained on only the 300 hour Switchboard conversational
telephone speech dataset and trained on both Switchboard (SWB) and Fisher
(FSH)~\cite{Cieri2004Fisher}, a 2000 hour corpus collected in a similar manner
as Switchboard.  Many researchers evaluate models trained only with 300 hours
from Switchboard conversational telephone speech when testing on Hub5'00.  In
part this is because training on the full 2000 hour Fisher corpus is
computationally difficult.  Using the techniques mentioned in
Section~\ref{section:optimization} our system is able perform a full pass over
the 2300 hours of data in just a few hours.

Since the Switchboard and Fisher corpora are distributed at a sample rate of
8kHz, we compute spectrograms of 80 linearly spaced log filter banks and an
energy term.  The filter banks are computed over windows of 20ms strided by
10ms.  We did not evaluate more sophisticated features such as the mel-scale
log filter banks or the mel-frequency cepstral coefficients.

Speaker adaptation is critical to the success of current ASR
systems~\cite{Povey2013,sainath2013cnn}, particularly when trained on 300 hour
Switchboard.  For the models we test on Hub5'00, we apply a simple form of
speaker adaptation by normalizing the spectral features on a per speaker basis.
Other than this, we do not modify the input features in any way.  

For decoding, we use a 4-gram language model with a 30,000 word vocabulary
trained on the Fisher and Switchboard transcriptions.  Again, hyperparameters
for the decoding objective are chosen via cross-validation on a held-out
development set.

The Deep Speech SWB model is a network of 5 hidden layers each with 2048
neurons trained on only 300 hour switchboard.  The Deep Speech SWB + FSH model
is an ensemble of 4 RNNs each with 5 hidden layers of 2304 neurons trained on
the full 2300 hour combined corpus.  All networks are trained on inputs of +/-
9 frames of context.

We report results in Table~\ref{table:hub5}. The model from Vesely et al.
(DNN-GMM sMBR)~\cite{Povey2013} uses a sequence based loss function on top of a
DNN after using a typical hybrid DNN-HMM system to realign the training set.
The performance of this model on the combined Hub5'00 test set is the best
previously published result.  When trained on the combined 2300 hours of data
the Deep Speech system improves upon this baseline by 2.4\% absolute WER and
13.0\% relative.  The model from Maas et al. (DNN-HMM
FSH)~\cite{maas2014largeam} achieves 19.9\% WER when trained on the Fisher 2000
hour corpus.  That system was built using Kaldi~\cite{Povey2011},
state-of-the-art open source speech recognition software.  We include this
result to demonstrate that Deep Speech, when trained on a comparable amount of
data is competitive with the best existing ASR systems.

\begin{table}[ht!]
\centering
\begin{tabular}{l  c  c  c }
\toprule
Model & SWB & CH & Full \\
\midrule
Vesely et al. (GMM-HMM BMMI)~\cite{Povey2013}   & 18.6 & 33.0 & 25.8 \\
Vesely et al. (DNN-HMM sMBR)~\cite{Povey2013}    & 12.6 & 24.1  & 18.4 \\
Maas et al. (DNN-HMM SWB)~\cite{maas2014largeam}  & 14.6 & 26.3  & 20.5 \\
Maas et al. (DNN-HMM FSH)~\cite{maas2014largeam}  & 16.0 & 23.7  & 19.9 \\
Seide et al. (CD-DNN)~\cite{seide2011}     & 16.1 & n/a & n/a \\
Kingsbury et al. (DNN-HMM sMBR HF)~\cite{Kingsbury2012}  & 13.3 & n/a & n/a \\
Sainath et al. (CNN-HMM)~\cite{sainath2013cnn} & 11.5 & n/a & n/a \\
Soltau et al. (MLP/CNN+I-Vector)~\cite{soltau2014cnn} & {\bf 10.4 } & n/a & n/a \\
{\bf Deep Speech SWB} & 20.0 & 31.8 & 25.9 \\
{\bf Deep Speech SWB + FSH} & 12.6 & {\bf 19.3} & {\bf 16.0} \\
\bottomrule
\end{tabular}
\caption{Published error rates (\%WER) on Switchboard dataset splits. The columns labeled ``SWB'' and ``CH'' are respectively the easy and hard subsets of Hub5'00.}
\label{table:hub5}
\end{table}

\subsection{Noisy speech}
\label{section:expnoise}

Few standards exist for testing noisy speech performance, so we constructed our
own evaluation set of 100 noisy and 100 noise-free utterances from 10 speakers.
The noise environments included a background radio or TV; washing dishes in a
sink; a crowded cafeteria; a restaurant; and inside a car driving in the rain.
The utterance text came primarily from web search queries and text messages, as
well as news clippings, phone conversations, Internet comments, public
speeches, and movie scripts. We did not have precise control over the
signal-to-noise ratio (SNR) of the noisy samples, but we aimed for an SNR
between 2 and 6 dB. 

For the following experiments, we train our RNNs on all the datasets (more than
7000 hours) listed in Table~\ref{table:datasets}.  Since we train for 15 to 20
epochs with newly synthesized noise in each pass, our model learns from over
100,000 hours of novel data.  We use an ensemble of 6 networks each with 5
hidden layers of 2560 neurons. No form of speaker adaptation is applied to the
training or evaluation sets.  We normalize training examples on a per utterance
basis in order to make the total power of each example consistent.  The
features are 160 linearly spaced log filter banks computed over windows of 20ms
strided by 10ms and an energy term.  Audio files are resampled to 16kHz prior
to the featurization. Finally, from each frequency bin we remove the global
mean over the training set and divide by the global standard deviation,
primarily so the inputs are well scaled during the early stages of training.

As described in Section~\ref{section:languagemodel}, we use a 5-gram language
model for the decoding.  We train the language model on 220 million phrases of
the Common Crawl\footnote{commoncrawl.org}, selected such that at least 95\% of
the characters of each phrase are in the alphabet. Only the most common 495,000
words are kept, the rest remapped to an \texttt{UNKNOWN} token.

We compared the Deep Speech system to several commercial speech systems: (1)
wit.ai, (2) Google Speech API, (3) Bing Speech and (4) Apple
Dictation.\footnote{wit.ai and Google Speech each have HTTP-based APIs. To test
Apple Dictation and Bing Speech, we used a kernel extension to loop audio
output back to audio input in conjunction with the OS X Dictation service and
the Windows 8 Bing speech recognition API.} 

Our test is designed to benchmark performance in noisy environments.  This
situation creates challenges for evaluating the web speech APIs:  these systems
will give no result at all when the SNR is too low or in some cases when the
utterance is too long.  Therefore we restrict our comparison to the subset of
utterances for which all systems returned a non-empty result.\footnote{This
leads to much higher accuracies than would be reported if we attributed 100\%
error in cases where an API failed to respond.} The results of evaluating each
system on our test files appear in Table~\ref{table:originalaudio}.  

To evaluate the efficacy of the noise synthesis techniques described in
Section~\ref{section:noisesynth}, we trained two RNNs, one on 5000 hours of raw
data and the other trained on the same 5000 hours plus noise.  On the 100 clean
utterances both models perform about the same, 9.2\% WER and 9.0\% WER for the
clean trained model and the noise trained model respectively.  However, on the
100 noisy utterances the noisy model achieves 22.6\% WER over the clean model's
28.7\% WER, a 6.1\% absolute and 21.3\% relative improvement.

\begin{table}[ht!]
\centering
\begin{tabular}{l  c  c  c}
\toprule
System      &  Clean (94) &  Noisy (82) & Combined (176) \\
\midrule
Apple Dictation  & 14.24   & 43.76  & 26.73 \\
Bing Speech      &  11.73     &   36.12   &  22.05   \\
Google API       &  6.64   & 30.47  & 16.72 \\
wit.ai           &  7.94   & 35.06  & 19.41 \\
{\bf Deep Speech}       &  {\bf 6.56}   & {\bf 19.06}  & {\bf 11.85} \\
\bottomrule
\end{tabular}
\caption{Results (\%WER) for 5 systems evaluated on the original audio. Scores are reported {\it only} for utterances with predictions given by all systems. The number in parentheses next to each dataset, e.g. Clean (94), is the number of utterances scored.}
\label{table:originalaudio}
\end{table}

\section{Related Work}
\label{section:related}

Several parts of our work are inspired by previous results. Neural network
acoustic models and other connectionist approaches were first introduced to
speech pipelines in the early 1990s~\cite{Bourlard93, Renals1994, Ellis1999}.
These systems, similar to DNN acoustic
models~\cite{Mohamed2011,Hinton2012,Dahl2011a}, replace only one stage of the
speech recognition pipeline.  Mechanically, our system is similar to other
efforts to build end-to-end speech systems from deep learning algorithms.  For
example, Graves~et~al.~\cite{Graves2006} have previously introduced the
``Connectionist Temporal Classification'' (CTC) loss function for scoring
transcriptions produced by RNNs and, with LSTM networks, have previously
applied this approach to speech~\cite{Graves2014}.  We similarly adopt the CTC
loss for part of our training procedure but use much simpler recurrent networks
with rectified-linear
activations~\cite{glorot2011deep,Maas2013,nair2010relurbm}.   Our recurrent
network is similar to the bidirectional RNN used by Hannun et
al.~\cite{hannun2014firstpass}, but with multiple changes to enhance its
scalability. By focusing on scalability, we have shown that these simpler
networks can be effective even without the more complex LSTM machinery.

Our work is certainly not the first to exploit scalability to improve
performance of DL algorithms.  The value of scalability in deep learning is
well-studied~\cite{coates2011kmeans,le2012faces} and the use of parallel
processors (including GPUs) has been instrumental to recent large-scale DL
results~\cite{GoogLeNet,le2012faces}.  Early ports of DL algorithms to GPUs
revealed significant speed gains~\cite{raina2009large}.  Researchers have also
begun choosing designs that map well to GPU hardware to gain even more
efficiency, including
convolutional~\cite{krizhevsky2012imagenet,ciresan2011highperf,Sainath2013} and
locally connected~\cite{coates2013cotshpc,ciresan2012multicolumn} networks,
especially when optimized libraries like cuDNN~\cite{Chetlur:14} and BLAS are
available.  Indeed, using high-performance computing infrastructure, it is
possible today to train neural networks with more than 10 billion
connections~\cite{coates2013cotshpc} using clusters of GPUs.  These results
inspired us to focus first on making scalable design choices to efficiently
utilize many GPUs before trying to engineer the algorithms and models
themselves.

With the potential to train large models, there is a need for large training
sets as well.  In other fields, such as computer vision, large labeled training
sets have enabled significant leaps in performance as they are used to feed
larger and larger DL systems~\cite{GoogLeNet,krizhevsky2012imagenet}.  In
speech recognition, however, such large training sets are less common, with
typical benchmarks having training sets ranging from tens of hours (e.g. the
Wall Street Journal corpus with 80 hours) to several hundreds of hours (e.g.
Switchboard and Broadcast News). Larger benchmark datasets, such as the Fisher
corpus~\cite{Cieri2004Fisher} with 2000 hours of transcribed speech, are rare
and only recently being studied.  To fully utilize the expressive power of the
recurrent networks available to us, we rely not only on large sets of labeled
utterances, but also on synthesis techniques to generate novel examples.  This
approach is well known in computer
vision~\cite{sapp2008synth,lecun2004learningmethods,coates2011icdar} but we
have found this especially convenient and effective for speech when done
properly.

\section{Conclusion}
We have presented an end-to-end deep learning-based speech system capable of
outperforming existing state-of-the-art recognition pipelines in two
challenging scenarios: clear, conversational speech and speech in noisy
environments.  Our approach is enabled particularly by multi-GPU training and
by data collection and synthesis strategies to build large training sets
exhibiting the distortions our system must handle (such as background noise and
Lombard effect).  Combined, these solutions enable us to build a data-driven
speech system that is at once better performing than existing methods while no
longer relying on the complex processing stages that had stymied further
progress.  We believe this approach will continue to improve as we capitalize
on increased computing power and dataset sizes in the future.

\section*{Acknowledgments} 
We are grateful to Jia Lei, whose work on DL for speech at Baidu has spurred us
forward, for his advice and support throughout this project.  We also thank Ian
Lane, Dan Povey, Dan Jurafsky, Dario Amodei, Andrew Maas, Calisa Cole and Li
Wei for helpful conversations.

\bibliography{references}
\bibliographystyle{abbrv}
\end{document}